\documentclass[10pt,twocolumn,letterpaper]{article}

\usepackage{cvpr}
\usepackage{times}
\usepackage{epsfig}
\usepackage{graphicx}
\usepackage{subfigure}

\usepackage{amsmath,amssymb,amsfonts}
\usepackage{nicefrac}       
\usepackage{booktabs}       
\usepackage{xcolor}
\usepackage{multirow}
\usepackage{caption}
\graphicspath{{imgs/}}

\usepackage[normalem]{ulem}

\usepackage[pagebackref=true,breaklinks=true,letterpaper=true,colorlinks,bookmarks=false]{hyperref}

\cvprfinalcopy 

\def\cvprPaperID{4255} 

\ifcvprfinal\fi
\begin{document}

\title{Learning Visual Knowledge Memory Networks for Visual Question Answering}

\author{Zhou Su$^{1}\thanks{This work was done when Zhou Su worked at Intel Labs China, and
Yinpeng Dong was intern at Intel Labs China. Jianguo Li is the corresponding author.}$,~ Chen Zhu$^2$,~ Yinpeng Dong$^3$,~ Dongqi Cai$^4$,~ Yurong Chen$^4$,~ Jianguo Li$^4$\\
$^1$Tencent Wechat,  $^2$ShanghaiTech University, $^3$Tsinghua University, $^4$Intel Labs China\\
{\tt\small zhousu@tencent.com, zhuchen@shanghaitech.edu.cn, dyp17@mails.tsinghua.edu.cn}\\
{\tt\small \{dongqi.cai,yurong.chen,jianguo.li\}@intel.com}}

\maketitle

\begin{abstract}
      Visual question answering (VQA) requires joint comprehension of images and natural language questions, where
    many questions can't be directly or clearly answered from visual content but require reasoning from structured human knowledge with confirmation from visual content.
    This paper proposes visual knowledge memory network (VKMN) to address this issue, which seamlessly incorporates structured human knowledge and deep visual features into memory networks in an end-to-end learning framework.
    Comparing to existing methods for leveraging external knowledge for supporting VQA, this paper stresses more on two missing mechanisms.
    First is the mechanism for integrating visual contents with knowledge facts.
    VKMN handles this issue by embedding knowledge triples (subject, relation, target) and deep visual features jointly into the visual knowledge features.
    Second is the mechanism for handling multiple knowledge facts expanding from question and answer pairs.
    VKMN stores joint embedding using key-value pair structure in the memory networks so that it is easy to handle multiple facts.
    Experiments show that the proposed method achieves promising results on both VQA v1.0 and v2.0 benchmarks, while outperforms state-of-the-art methods on the knowledge-reasoning related questions.
\end{abstract}

\section{Introduction}
\begin{figure}[]
\centering
\subfigure[]{\label{fig:1a}\includegraphics[height=0.295\linewidth]{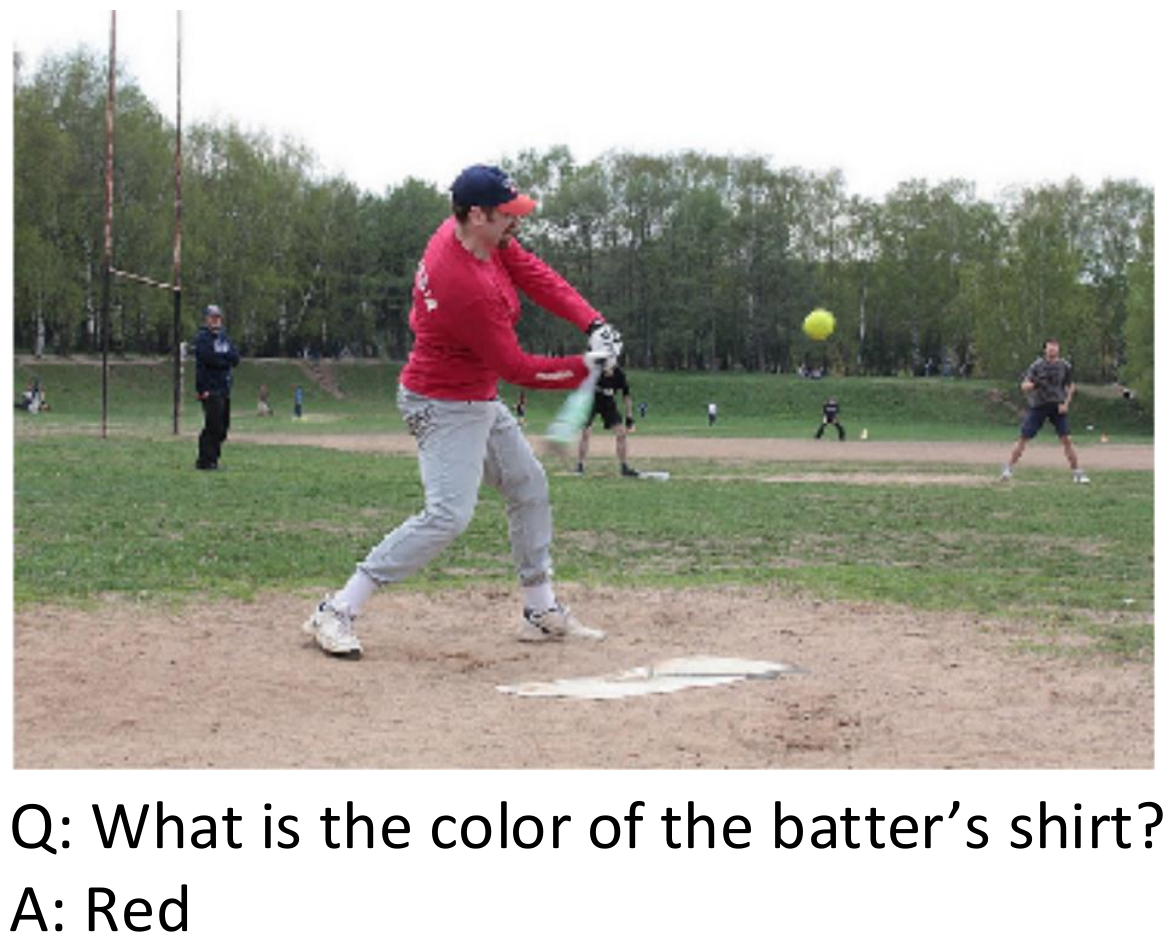}}
\hspace{-1ex}
\subfigure[]{\label{fig:1b}\includegraphics[height=0.295\linewidth]{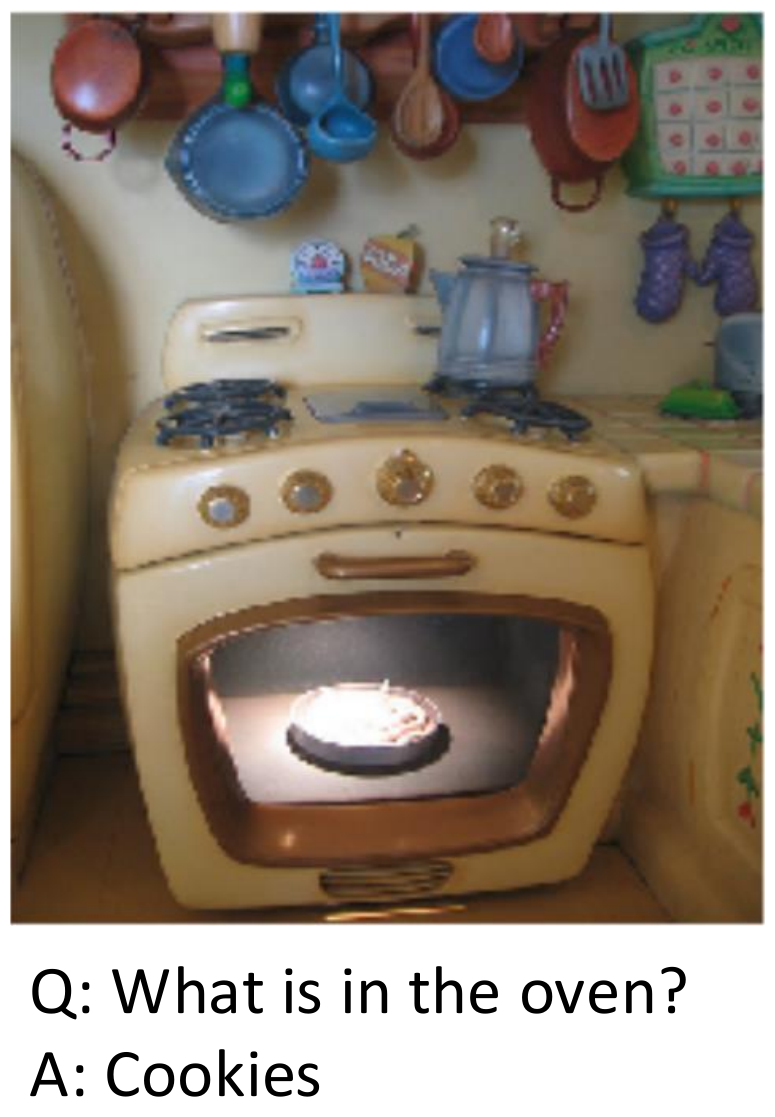}}
\hspace{-1ex}
\subfigure[]{\label{fig:1c}\includegraphics[height=0.295\linewidth]{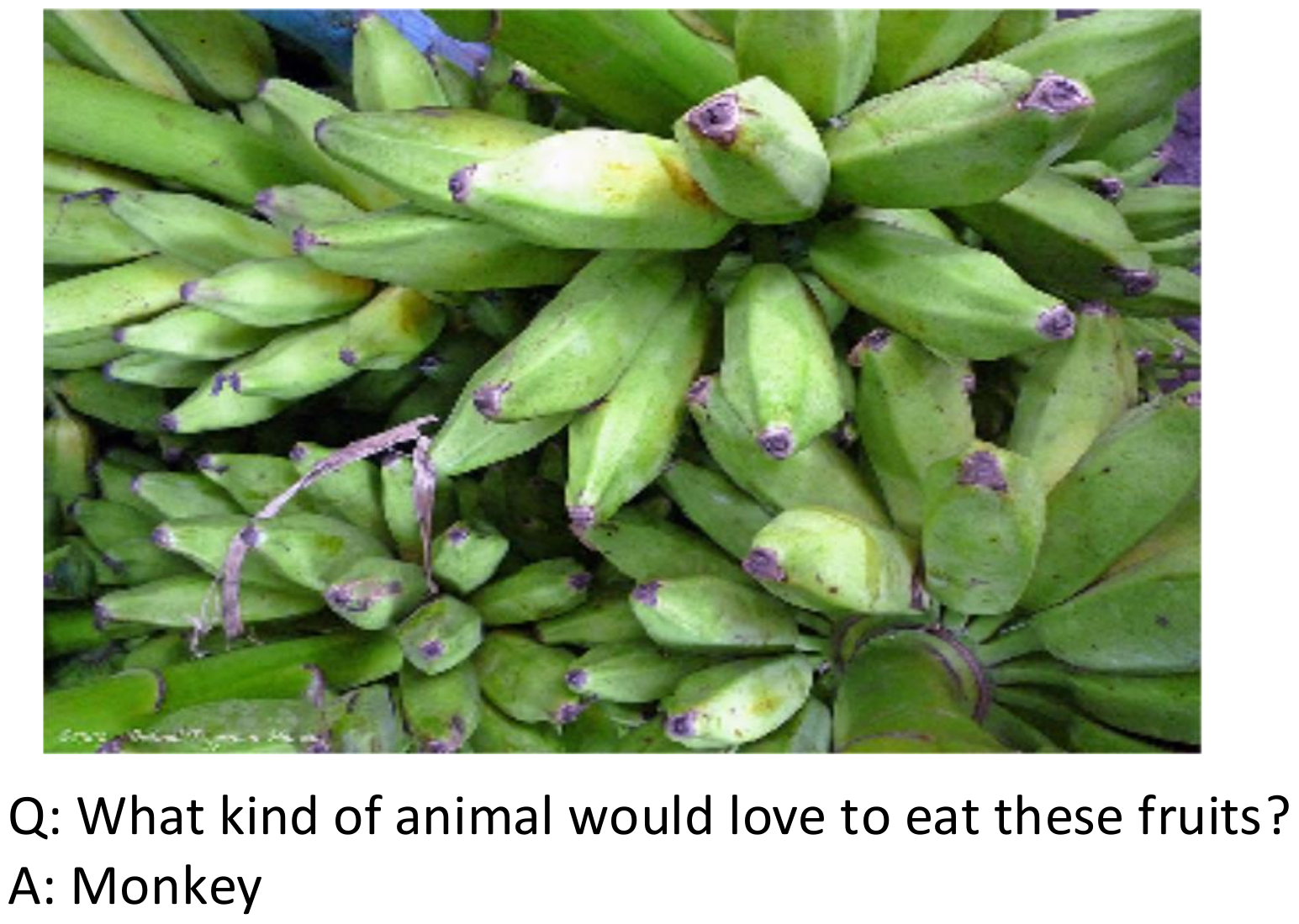}}
\vspace{-2ex}
\caption{Examples of different question objective categories in VQA. (a) is an example of \textit{apparent objective} that the answer is clearly visible and can be easily answered using recognition results.
(b) is an example of \textit{indiscernible objective} that the answer is unclear for visual recognition, and may need constraints from common sense. (c) is an example of of \textit{invisible objective} that the answer requires deduction/reasoning from external knowledge.}
\label{fig:example}
\vspace{-3ex}
\end{figure}

Visual Question Answering (VQA) is an emerging interdisciplinary research field in the last few years, which attracts broad attention from both computer vision and natural language processing (NLP) communities.
VQA aims to help computer automatically answer natural language question about an image.
The question answers can be divided into the following types: yes/no answers, multi-choice answers, numeric answers, and open-ended word/phrase answers (for questions about \textit{what,where,who,...}).
VQA requires comprehensive image and language understanding, which constitutes a truly AI-complete task with similar spirit as Turing test \cite{geman2015visual,malinowski2014towards,malinowski2015ask}.

Basically, VQA is formulated as a classification problem in most researches, in which images and questions are input, with answers as output categories (due to a limited number of possible answers).
As the VQA task was proposed after deep learning approaches had already gained wide popularity, almost all current VQA solutions use CNN to model image input and recurrent neural network (RNN) to model the question \cite{Zhou2015}.
Attention mechanism has been heavily investigated in VQA.
This includes visual attention~\cite{yang2016stacked,xu2016ask,shih2016look} which focuses on handling the problem {\textit{where to look}}, and question attention \cite{rocktaschel2015reasoning,yin2015abcnn,dos2016attentive,lu2016hierarchical}
which focuses on solving the problem {\textit{where to read}}.
As images and questions are two different modalities, it is straightforward to jointly embed two modalities together for a unified description of the image/question pair. Some works \cite{fukui2016multimodal,Kima,kim2016hadamard,Zhu_2017_ICCV,Yu_2017_ICCV} even consider putting the attention mechanism and multi-modal joint embedding in one unified framework.

However, VQA is significantly more complex than other vision and language tasks such as image captioning, since explicit information (middle-level recognition results such as objects, attributes, or even image captions, etc) are not enough for accurate VQA.
For instance, we investigated the VQA v1.0 dataset \cite{Agrawal2015}, which is based on the COCO dataset \cite{Lin2014}.
As each image has 5 manual caption annotations, we match question answers to words in image captions, which only produces 2,907 exact matches,
while the remaining nearly 50K answers do not appear in captions. Basically, we could divide the question objectives of VQA into three categories:
(a) \textit{Apparent objective} which answers in the query image could be directly obtained from recognition results (objects, attributes, captions, etc);
(b) \textit{Indiscernible objective} which answer targets are usually too small or unclear in the query image, and thus requires supporting facts for correct answers;
(c) \textit{Invisible objective} which requires deduction of common sense, topic-specific or even encyclopedic knowledge about the content of the image.
Please refer to \autoref{fig:example} for some examples of these three cases.
The external knowledge information at least could help VQA on the latter two categories.
A supporting data is that we find more than 49,866 answers in VQA v1.0 dataset appear in knowledge base from the visual genome \cite{Krishna2016}.
A few pioneering works~\cite{wang2015explicit,wang2016fvqa,wu2016ask} study the problem on \textit{how to reason} with prior knowledge information for the VQA task.
They only involve one supporting fact to help the decision, which may introduce knowledge ambiguity/inaccuracy due to the inaccurate knowledge extraction procedure, and further yield wrongly question answering.

Before the VQA task appears, NLP community has extensively studied the text only question-answering (QA) problem \cite{voorhees2000building,kolomiyets2011survey}.
Classical methods read documents directly and use information retrieval methods to find answers \cite{kolomiyets2011survey}.
Thereafter, knowledge base (KB) such as Freebase \cite{bollacker2008freebase} organizes information into structured triples:
\textit{\textless $s$, $r$, $t$ \textgreater}, where $s$ is the subject, $t$ is the target, and $r$ is the relation between $s$ and $t$.
Then, question answering is converted into a database query problem \cite{berant2013semantic,fader2014open}.
Most recently, memory networks have been proposed to combine document reading and knowledge base retrieval \cite{weston2014memory,Miller,sukhbaatar2015end}
for accurate QA.

\begin{figure*}[]
\centering
\includegraphics[width=0.8\linewidth]{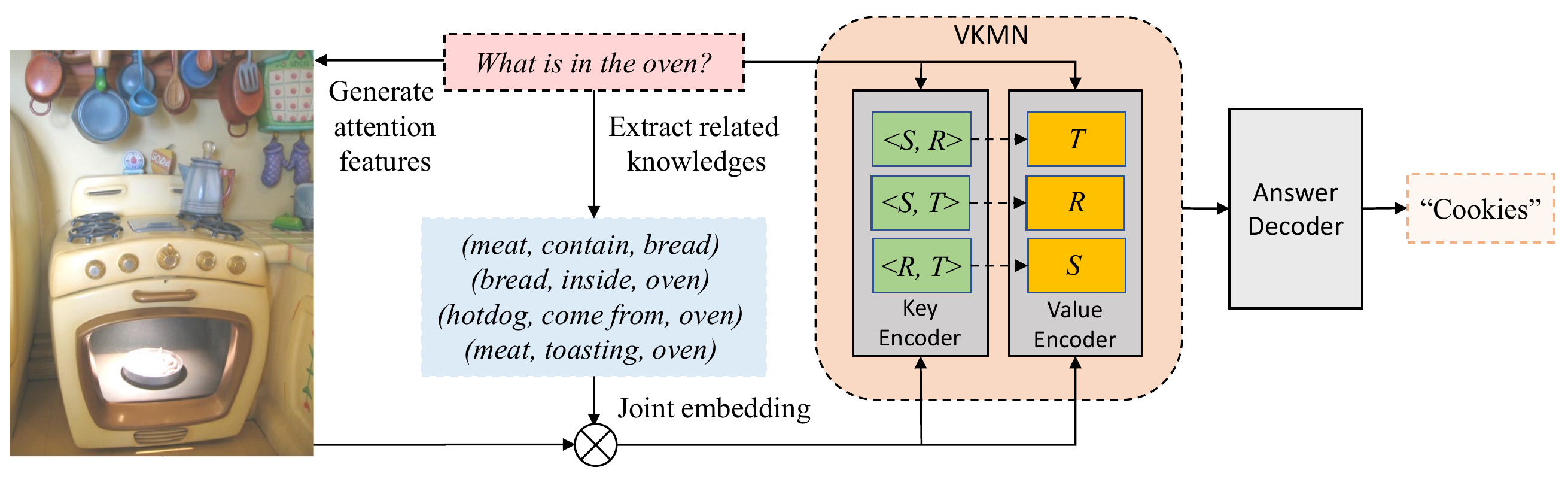}
\vspace{-1ex}
\caption{Illustration of VKMN for the VQA task.
    Note that three replicated memory sub-blocks (different combination of $s$, $r$, $t$ as key-part or value-part) are used to handle the ambiguity on which part of the knowledge triple is missing in the query question.}
\label{fig:demo}
\vspace{-2ex}
\end{figure*}

Inspired by the development of memory networks based text QA methods, this paper proposes visual knowledge memory network (VKMN) for
accurate \textit{reasoning} with a pre-built visual knowledge base. \autoref{fig:demo} illustrates how the proposed VKMN model works on visual question answering.
VKMN extracts multiple related knowledge facts from the question, jointly embeds knowledge facts with visual attentive features into visual knowledge attentive features, and stores them in a key-value pair for easy and efficient reading from memory.
The memory reading will make the visual-question addend to the highly relevant/correlated knowledge facts, and thus gives much more accurate question answering.
The major contributions of this paper are:
\begin{itemize}
\setlength{\topsep}{1pt}
\setlength{\itemsep}{1pt}
\setlength{\parskip}{1pt}
\item[(1)] We propose VKMN, a simple yet efficient end-to-end trainable framework, which inherits the merits from attention based methods and joint-embedding based methods, while avoids the knowledge inaccuracy limitation of current knowledge-based solutions.
\item[(2)] We build a visual-question specific knowledge base, which does not contain irrelevant knowledge entries as generic knowledge base like Freebase \cite{bollacker2008freebase}.
\item[(3)] We conduct extensive experiments on the VQA v1.0 and v2.0 benchmark datasets, and show that the proposed method achieves competitive accuracy, while outperforms state-of-the-art methods on knowledge-reasoning related questions.
\end{itemize}

\section{Related Work}
\noindent{\textbf{Attention mechanisms in VQA.}}
Attention Mechanisms in Neural Networks are loosely based on the visual attention mechanism found in human brains.
It was first proposed and successfully applied in tasks like machine translation \cite{bahdanau2014neural}, image captioning \cite{xu2015show},
while then became popular in VQA.
The main idea is that specific parts of the input (image and/or question) are more informative/effective than others for answering a given question.
Numerous VQA methods \cite{xu2016ask,shih2016look,lu2016hierarchical,yang2016stacked,andreas2016NMN} have incorporated spatial attention to learn specific CNN features according to the input question, rather than using holistic global features from the entire image. Some have also incorporated attention into the text representation.
For instance, recent works~\cite{lu2016hierarchical} focus on the co-attention models that jointly exploit visual attention and question attention advantages through a unified hierarchical architecture.
Furthermore, there are some researches combining attention mechanism with multi-modal representation through joint embedding visual attention features and text question features.
This includes some state-of-the-art VQA methods: Multimodal Compact Bilinear pooling (MCB)~\cite{fukui2016multimodal}, Multimodal Low-rank Bilinear pooling (MLB)~\cite{kim2016hadamard}, Multi-modal Factorized Bilinear pooling (MFB) \cite{Yu_2017_ICCV}, and Structured Visual Attention (SVA) \cite{Zhu_2017_ICCV}.

This paper does not study the spatial attention mechanisms, while directly leverage existing multi-modal attention method as one component in our method.

\noindent{\textbf{Knowledge base and VQA.}}
Knowledge base (KB) is a technology used to store structured fact entries in the triple form
\textit{\textless $s$, $r$, $t$ \textgreater}, where $s$ is the subject, $t$ is the target, and $r$ is the relation between $s$ and $t$.
There is an increasing interest in the NLP community using KBs for question answering \cite{berant2013semantic,fader2014open}.
However, there are still limited works on leveraging KBs for VQA.
As is known, VQA is an inter-discipline task in vision and language just like image/video captioning \cite{xu2015show,fang2015captions,shen2017weakly},
but moves beyond for even deeper image understanding,
since it often requires information not contained in the image itself, which can range from ``common sense" knowledge to some topic-specific knowledge.
For instance, understanding the image content is not enough to answer the question in Figure~\autoref{fig:1c}.
The VQA system must first recognize that the ``fruit" entity is ``banana", and inference based on knowledge about ``animal loving banana" to get the answer ``monkey".
We also argue that for some cases such as Figure~\autoref{fig:1b}, even if the target content is visible, it may be too small and/or unclear to yield
a wrong recognition result. KB will then act as an probabilistic prior to adjust the decision score on a list of candidates and output a correct
result. We will verify these two cases with some qualitative examples in our experiments.

Substantial studies have focused on building large-scale structured Knowledge Bases (KBs) for QA, such as DBpedia~\cite{auer2007dbpedia}, Freebase~\cite{bollacker2008freebase}, ConcepNet~\cite{liu2004conceptnet}, etc.
Zhu \textit{et al.} \cite{zhu2015VKB} even built large-scale multimodal knowledge base for VQA purpose.
A few works~\cite{wang2015explicit,wang2016fvqa,wu2016ask} also tried to introduced KBs in the VQA task.
Wang {\em et al.}~\cite{wang2015explicit} proposes ``Ahab" method to reason about the content of an image, which first detects concepts in the query image and then links them to the relevant parts of DBpedia KB. It learns the mapping of images/questions to queries over the constructed knowledge graph for final question answering.
This method is restricted to questions parsed with manually designed templates.
\cite{wang2016fvqa, wu2016ask} improve ``Ahab" by introducing long-short term memory (LSTM) and a data-driven approach to learn the mapping of images/questions to queries with the knowledge fact meeting the search conditions in a KB.

All these methods only involve one supporting knowledge fact to help the decision, so that they suffer greatly from the knowledge inaccuracy problem due to the inaccurate extraction procedure. This paper tries to alleviate this issue with memory network mechanism to handle multiple knowledge facts expanding from questions.

\noindent{\textbf{Memory networks.}}
Memory networks were first proposed for modern large-scale question answering (QA) systems in~\cite{weston2014memory},
which introduces a long-term memory block for reading/writing simple facts when reasoning with inference components in the QA task.
Sukhbaatar {\em et al.}~\cite{sukhbaatar2015end} improve memory networks with an end-to-end learning fashion, which requires fewer supervision signals during training stage and has more practicality.
To bridge the gap between KB query/inference and documents reading, Miller {\em et al.}~\cite{Miller} propose Key-Value Memory Networks (KV-MemNN) which can read documents and answer questions more viable and effective. The KV-MemNN performs QA by first storing facts in a key-value structured memory before reasoning over them for the answer.

Memory networks were later introduced into the VQA task in the form of dynamic memory networks \cite{xiong2016dynamic}, which consists of four modules:
input module for encoding input with a set of ``facts" vectors, question module for deriving a vector representation of the question,
an episodic memory module for retrieval attention facts to the question, and answer module for combining final memory state and the
question vector to predict the output.
Similarly, spatial memory networks \cite{xu2016ask} also store CNN features from image grid regions/patches into memory,
and selects certain parts of the information with an explicit attention mechanism for question answering.

We also try to utilize memory network for VQA. We have at least two differences to \cite{xiong2016dynamic,xu2016ask}. First, we design a key-value memory network for visual knowledge attentive features rather than a simple dynamic memory network as in \cite{xiong2016dynamic,xu2016ask}. Second, the memory network we used focuses more on multiple visual knowledge encoding, while \cite{xu2016ask} focuses more on spatial patches encoding.

\begin{figure*}[]
\centering
\includegraphics[width=0.8\linewidth]{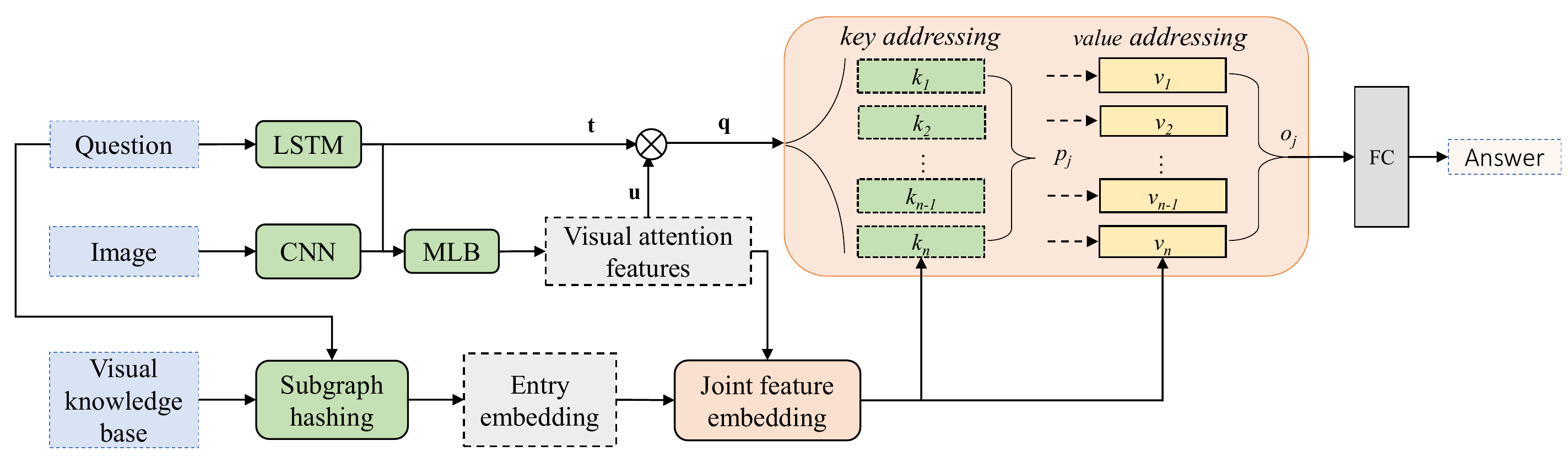}
\vspace{-2ex}
\caption{Diagram of Visual Knowledge Memory Network based VQA system}
\label{fig:diagram}
\vspace{-3ex}
\end{figure*}

\section{Method}
\subsection{Overview of the Method}
We propose the visual knowledge memory network (VKMN) aiming to answer the visual question more accurately with an auxiliary visual knowledge base.
VKMN is derived from key-value memory network \cite{Miller}, which has been proven effective in the QA task.
VKMN constructs a visual-question related knowledge base in advance, which allows great flexibility for designing key and value items based on the visual features and (part-of) knowledge triples.
Basically, our VKMN model consists of four modules:
\begin{itemize}
\setlength{\topsep}{1pt}
\setlength{\itemsep}{1pt}
\setlength{\parskip}{1pt}
\item[(1)] \textit{Input module}, which encodes input images with CNN model and questions with a RNN model, and further obtains the query representation by jointly embedding the output of these two models;
\item[(2)] \textit{Knowledge spotting module}, which retrieves related knowledge entries based on query questions or automatic image captions from the pre-built visual knowledge base by sub-graph hashing;
\item[(3)] \textit{Joint visual and knowledge embedding module}, which embeds visual features and knowledge triples (or part of the triples) jointly for the ease of storing in key-value memory networks;
\item[(4)] \textit{Memory module}, which receives query question, reads the stored key-value to form the visual-knowledge feature, and predicts the answers.
\end{itemize}
\autoref{fig:demo} illustrates how the proposed VKMN model works on visual question answering, and \autoref{fig:diagram} gives detailed diagram on how different modules interact in VKMN. We will elaborate each module separately below.

\subsection{Encoding of Image/Question Inputs}
The input image $I$ and question $q$ need to be processed into feature vectors before feeding into memory networks.
We process input images with an ImageNet pre-trained CNN model, and questions with a LSTM model.
The feature vectors from two modalities could be jointly embedded into a single visual attentive description as defined in Eq.\ref{eq:vad} for answer prediction.
Several methods are proposed to learn the multimodal joint embedding in an end-to-end manner for VQA, including the VQA 2016 challenge winner solution MCB \cite{fukui2016multimodal}, and the state-of-the-art solution MLB \cite{kim2016hadamard}.

In this paper, we directly leverage MLB for visual-question pair encoding. We denote the MLB with spatial attention output (aka the first MLB stage output) as $\mathbf{u}$, and the LSTM encoded question vector as $\mathbf{t}$, where $\mathbf{t}$ is already projected to the same dimensional space as $\mathbf{u}$ with one internal fully-connection (FC) layer, i.e., $\mathbf{t}, \mathbf{u} \in \mathcal{R}^d$. The query representation is the joint embedding of $\mathbf{t}$ and $\mathbf{u}$ with low-rank bilinear pooling \cite{kim2016hadamard} as
\begin{equation}
\label{eq:vad}
    \mathbf{q} = \mathbf{t} \odot \mathbf{u},
\end{equation}
where $\odot$ indicates Hadamard product between two vectors (element-wise product).
$\mathbf{q}$ is the visual attentive description of the visual-question pair for query purpose.

\subsection{Knowledge Spotting with Sub-graph Hashing}
\begin{figure}[]
   		\centering
   		\includegraphics[width=0.99\linewidth]{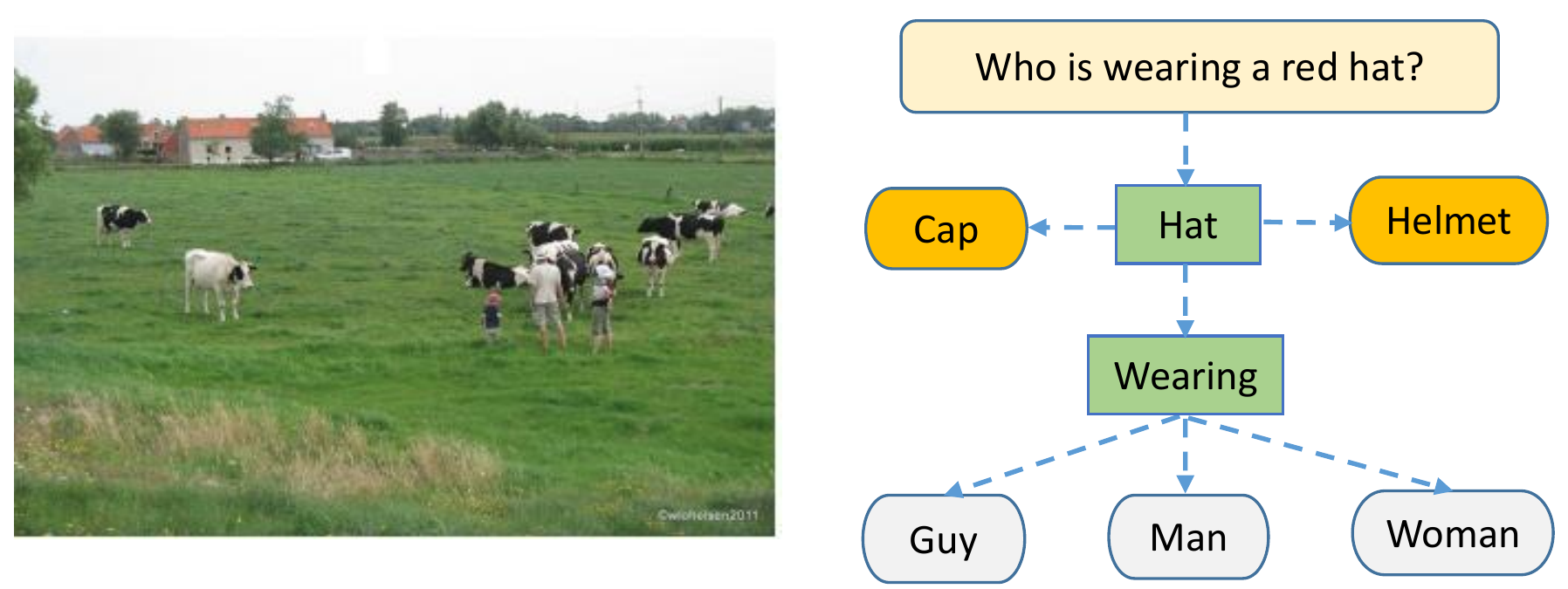}
   		\vspace{-1ex}
   		\caption{Illustration of knowledge subgraph expansion. We do question sentence parsing to get the target `hat' and relation `wearing'.
        We expand the subject with query from knowledge-base (results in yellow), and also expand the target (results in gray) based on subject plus relation.
    Dashed-line indicates expansion results.}
   		\label{subgraph} 
   		\vspace{-3ex}
   	\end{figure}

Before elaborating the details of visual knowledge representation, we present how to spot knowledge entries related to the visual question.
Firstly, given all the knowledge triples \textit{\textless $s_i$, $r_i$, $t_i$ \textgreater} in the pre-built Visual Knowledge Base,
we generate the entity set $E = \{s_i, t_i\}$, and relation set $R=\{r_i\}$.
We call $S = E\cup R$ the entry set, which contains all the different entries in the knowledge base.
Then we extract entries whenever one phrase in questions (or automatic generated captions) matches one item in the entry set $S$ using sub-graph hashing methods \cite{bordes2014subgraph}.
To alleviate the inaccuracy of extracted visual knowledge, we restrict that each knowledge triple should contain at least two entries extracted from a question (or automatic generated captions), which is more stringent than that in MemNN~\cite{weston2014memory, sukhbaatar2015end} and KV-MemNN~\cite{Miller}.
A small subset of $N$ knowledge triples \{\textless $s_{1}$, $r_{1}$, $t_{1}$\textgreater, $\cdots$, \textless $s_{N}$, $r_{N}$, $t_{N}$\textgreater\}  are created afterwards.
To handle the long-tail effects in the visual knowledge base,
we perform knowledge triple expansion on the knowledge graph to include the direct neighborhood of those extracted $N$ knowledge triples.
\autoref{subgraph} illustrates one example of subgraph hashing and knowledge triple expansion.
Finally, we setup a memory network which could store $M$ knowledge entries ($M>N$).
If the size of expanded knowledge subset is less than $M$, we append null entries.

\subsection{Joint Embedding of Visual and Knowledge}
In the QA task \cite{Miller}, people just take the knowledge triple or part-of-knowledge triple to compose the key-value pair in key-value memory networks.
In VQA, visual contents are essential to answering the visual questions.
Our start point is the spatial attentive visual feature $\mathbf{u}$ from the input module and knowledge entry $\mathbf{e}$ from knowledge spotting module.
We are required to learn a joint embedding to combine $\mathbf{u} \in \mathcal{R}^d$ and $\mathbf{e}$ together.
As $\mathbf{e}$ is text representation, we impose a mapping function $\Phi(\cdot)$ to get real-valued feature vector $\Phi(\mathbf{e}) \in \mathcal{R}^{d_e}$.
Here $\Phi(\cdot)$ could be either bag-of-words (BoW) representation, word2vec transformation, or even knowledge embedding like TransE \cite{bordes2013TransE}. The feature dimension of $\mathbf{u}$ and $\Phi(\mathbf{e})$ are usually different, i.e., $d \neq d_e$.
We project them into the same space, and apply MLB \cite{kim2016hadamard} to capture their joint representation as
\begin{equation}
\label{eq:embed}
  \mathbf{x} = \Psi(\mathbf{e}, \mathbf{u}) = \sigma(W_e \Phi(\mathbf{e}) )\odot\sigma(W_u \mathbf{u}),
\end{equation}
where $\sigma(\cdot)$ is hyperbolic tangent function (favors over sigmoid function in our experiments), $W_u$ and $W_e$ are the matrices to project $\mathbf{u}$ and $\Phi(\mathbf{e})$ into the same dimensional space.
$\mathbf{x}$ is called visual knowledge attentive description, since it integrates visual feature $\mathbf{u}$ with knowledge entry $\mathbf{e}$.

We argue that $\mathbf{x}$ captures more fine-grained information about the knowledge than the spatial attentive visual feature $\mathbf{u}$ from MLB, in which the attention is based on the whole question. We verified this point in the ablation study (see row 4 \& 6 of \autoref{tab:ablation}).

\subsection{Visual Knowledge Memory Network}\label{sec:vkmn}
We presented general joint embedding of visual features and knowledge entries in previous subsection.
As VKMN defines memory slots as key-value vector pairs like $(\mathbf{k}_1, \mathbf{v}_1),\cdots, (\mathbf{k}_M, \mathbf{v}_M)$, in this subsection,
we discuss the key-value pair design first, and the inference procedure with memory networks later.

\textbf{Key-value design.} The design of which part should be key, and which part should be value is quite flexible.
In traditional QA problem, given the knowledge triple \textit{\textless $s$, $r$, $t$ \textgreater}, people usually consider the first two ($s$ and $r$) as key, while take the last one $t$ as value. However, this is not true for VQA, as we do not know which part of the knowledge triple is missing in the visual question.
There are three combinations of keys while taking the remained item as value: (1) ($s$, $r$) as the key; (2) ($s$, $t$) as the key; (3) ($r$, $t$) as the key.
In practice, we build three memory blocks for these three cases separately as shown in \autoref{fig:demo}, and name it \textit{triple replication}.
This is useful to distinguish questions like ``what is the toothbrush used for?" and ``what is used for brushing teeth?".
For simplicity, we only elaborate the case ($s$, $r$) as the key item and $t$ as the value item in the following study.
We can employ Eq.\ref{eq:embed} to obtain the embedding of keys and values. Suppose $\mathbf{e}=(e_1, e_2, e_3)$, in which $e_1$, $e_2$ and $e_3$ correspond to $s$, $r$, $t$ according to the designed key-value mapping.
To ensure that key representation $\mathbf{k}_i$ and value representation $\mathbf{v}_i$ are of the same dimensionality,
we make the additive assumption similar to the continuous bag-of-words (CBOW), and derive the $\mathbf{k}_i$ and $\mathbf{v}_i$ as below:
\begin{eqnarray}
    \mathbf{k}_i &=& \Psi(e_1, \mathbf{u}_i) + \Psi(e_2, \mathbf{u}_i); \\
    \mathbf{v}_i &=& \Psi(e_3, \mathbf{u}_i).
\end{eqnarray}
This additive assumption also suits for TransE encoding \cite{bordes2013TransE},
since it adopts shared weights to encode the knowledge triple \textit{\textless $s$, $r$, $t$ \textgreater} as a whole, and outputs the same dimensional feature vector for $s$, $r$, and $t$ respectively.

With the designed key-value pairs stored in VKMN, the inference consists of three steps: addressing related knowledge, reading corresponding value and answering the question.
We discuss them step by step below.

\textbf{Key addressing.}
Given a query $\mathbf{q}$, we address each candidate memory slot and assign a relevance probability by comparing the question to each key:
\begin{equation}
\label{eq:score}
    p_i = Softmax( \mathbf{q} \cdot A \mathbf{k}_i),
\end{equation}
where $\cdot$ denotes inner product, $A$ is the parameter matrix for memory networks which projects $\mathbf{k}_i$ into the same feature space as $\mathbf{q}$,
and $Softmax(z_i) = \nicefrac{e^{z_i}}{\sum_j e^{z_j}}$.

\textbf{Value reading.}
In the value-reading step, the values of memory slots are read by weight averaging with the addressing possibilities, and the returned vector $\mathbf{o}$ is defined as:
\begin{equation}\label{eq:pav}
    \mathbf{o} =  \sum\nolimits_i p_i A \mathbf{v}_i
\end{equation}
The original key-value memory networks \cite{Miller} supports iterative refining the query and reading the memories.
In this paper, we only make one-step update of query with $\mathbf{q}' = \mathbf{q} + \mathbf{o}$ after receiving $\mathbf{o}$.

\textbf{Question answering.}
The number of different answers is fixed for the VQA task so that the question answering is re-casted to be a classification problem.
We predict the answer based on $\mathbf{q}'$ using a FC layer with weight matrix $W_o$ as below
\begin{equation}\label{eq:smax}
\hat{a} = \arg\max Softmax(W_o \mathbf{q}').
\end{equation}
All the parameter matrix $W_u$, $W_e$, $A$, and $W_o$ in VKMN are end-to-end trained with stochastic gradient descent (SGD) based backpropagation algorithm.

\section{Experiments}
We evaluate the proposed model on the VQA v1.0 \cite{Agrawal2015} and v2.0\cite{balanced_vqa_v2} datasets. We will elaborate the details of the dataset,
the procedure of building visual knowledge base, ablation studies on different configurations and the benchmark results below separately.

\subsection{Datasets}
The \textit{VQA v1.0} dataset \cite{Agrawal2015} consists of 204,721 MS-COCO images \cite{Lin2014}, with 614,163 associated question-answer pairs.
There are three designed data splits: train (248,349 questions), validation (121,512 questions) and test (244,302 questions).
In addition, 25\% subset of test partition is splitted as test-dev set.

The \textit{VQA v2.0} dataset \cite{balanced_vqa_v2} further balanced the v1.0 dataset by collecting complementary images such that
each question in the balanced dataset has a pair of similar images but with different  answers to  the question.
It finally consists of 1,105,904 questions from 204,721 images, which is about twice larger than v1.0 dataset.

For simplicity, we restricted the target answers to the top 2,000 most frequent answers in the train+val set as most existing works \cite{kim2016hadamard,fukui2016multimodal}.
Therefore, our training data consists of data with the top 2,000 answers in the train+val set, which covers about 90\% question-answer pairs in the train+val set.

The \textit{Visual Genome} dataset \cite{Krishna2016} contains 108,249 images labeled with question-answer pairs, objects and attributes. An average of 17 question-answer pairs are collected for each image. Moreover, there are \textit{Visual Genome Relationship} (VGR) annotations available.
\textit{VGR} provides relationship descriptions of different objects on images in the triple form \textit{\textless $s$, $r$, $t$\textgreater}, which perfectly fits the needs of our visual knowledge base.
Detailed usage settings are given in section \ref{Building Visual Knowledge Base}.

\subsection{Building Visual Knowledge Base}
\label{Building Visual Knowledge Base}
As the general purpose knowledge bases like Freebase contains billions of knowledge facts, while most knowledge entries are irrelevant to the visual questions,
we build our own knowledge base for the purpose of VQA, in which each entry has a structure of \textit{\textless $s$, $r$, $t$\textgreater}.
The entries in the knowledge base (named visual knowledge base) come from two sources: (i) knowledge entries extracted from the question-answer pairs in the VQA v1.0 train+val set \cite{Agrawal2015};
(ii) knowledge triples from the \textit{VGR} dataset~\cite{Krishna2016}.

First, we perform information extraction from the question-answer pairs into the VQA v1.0 train+val set, to obtain a bunch of structured knowledge entries.
Compared to existing knowledge bases like DBpedia \cite{auer2007dbpedia} and ConceptNet \cite{liu2004conceptnet},
knowledge entries we extracted are more closely bound up with questions associated with their images, and the expression is more precise and compact than original question-answer pairs in the VQA dataset \cite{Agrawal2015}.
The knowledge entry extraction procedure works as below.
We first parse the  part-of-speech (POS) tags and dependency tree of each question via the Stanford Parser \cite{Chen2014}.
Then we extract triples from QA pairs using question tags, dependencies, and answers.
All words in the triples are lemmatized so that inflectional variants of the same word explicitly share the same representation.
Finally, we build a \textit{relation set} $R=\{r\}$, where $r$ is all possible relation phrases in \textit{VGR} dataset\cite{Krishna2016}.
For each extracted triple \textit{\textless $s$, $r$, $t$\textgreater} from the VQA dataset, we replace $r$ with the most similar/closed one in set $R$.
Lemmatisation and relation replacement can help reducing data sparseness, especially for capturing ``long-tail" entries.

Second, the \textit{VGR} dataset \cite{Krishna2016} consists of 1,531,448 knowledge triples describing object relationships, about 14 triples per image.
We filter the original triple sets in VGR dataset to obtain 40,480 unique triples by removing
triples with appearance frequency (either $s$, $r$, or $t$) less than 3.

After combining these two parts together, we obtain a visual knowledge base with about 159,970 unique triples.

\begin{table}[]
  \centering
  \small
  \resizebox{\linewidth}{!}{
  \begin{tabular}{cccccc}
    \hline
    \multicolumn{1}{c}{Model} &\multicolumn{1}{c}{All} &\multicolumn{1}{c}{Y/N} &\multicolumn{1}{c}{Num} & \multicolumn{1}{c}{Other}\\
    \hline
     (1) BoW (GLoVe)         &65.5  &83.8 &38.4 &56.1   \\
     (2) Blind model (text only) &49.3   &77.9  &34.3  &28.3   \\
     (3) $\mathbf{q}$ only no KB (MLB) \cite{kim2016hadamard} & 65.1 & 84.1 &38.2 & 54.9  \\
     (4) No triple replication   &64.5  &84.0 &37.3 &53.9    \\
    \hline
    Baseline (full VKMN)              &66.0  &83.7 &37.9 &57.0\\
    \hline
  \end{tabular}
  }
  \vspace{-2ex}
  \caption{Ablation studies of the design choice on the \textit{Open-Ended} task of the VQA v1.0 test-dev dataset.}
  \label{tab:ablation}
\vspace{-3ex}
\end{table}

\begin{table*}[]
  \centering
  \small
  \resizebox{0.8\linewidth}{!}{
  \begin{tabular}{ccccccccccc}
    \toprule
    \multirow{3}*{Model} & \multicolumn{5}{c}{test-dev} & \multicolumn{5}{c}{test-standard} \\
    \cmidrule(lr){2-6}\cmidrule(lr){7-11}
    \multicolumn{1}{c}{} & \multicolumn{4}{c}{Open-Ended} & MC & \multicolumn{4}{c}{Open-Ended} & MC \\
    \cmidrule(lr){2-5}\cmidrule(lr){6-6}\cmidrule(lr){7-10}\cmidrule(lr){11-11}
    \multicolumn{1}{c}{} &\multicolumn{1}{c}{All} &\multicolumn{1}{c}{Y/N} &\multicolumn{1}{c}{Num} & \multicolumn{1}{c}{Other}  &\multicolumn{1}{c}{All} &\multicolumn{1}{c}{All} &\multicolumn{1}{c}{Y/N} &\multicolumn{1}{c}{Num} & \multicolumn{1}{c}{Other}  &\multicolumn{1}{c}{All}\\
    \midrule
    iBOWIMG\cite{Zhou2015} &55.7   &76.5 &35.0 &42.6 &-  &55.9  &76.78 &35.0 &42.6 &62.0    \\
    DPPnet\cite{Noh2015}   &57.2   &80.7 &37.2 &41.7 &-  &57.4  &80.23 &36.9 &42.2 &62.7   \\
    SMem\cite{xu2016ask}   &58.0   &80.9 &37.3 &43.1 &-  &58.2  &80.9  &37.5 &43.5 &-   \\
    AMA\cite{wu2016ask}   &59.2   &81.0 &38.4 &45.2 &-  &59.4  &81.1 &37.1 &45.8 &-   \\
    DMN+\cite{xiong2016dynamic} &60.3 &80.5 &36.8 &48.3 &- &- &- &- &- &-  \\
    MRN\cite{Kima}         &61.6   &82.2 &38.8 &49.2 &-  &61.8  &82.4 &38.2 &49.4 &66.3   \\
    HieCoAtt\cite{lu2016hierarchical}      &61.8   &79.7 &38.7 &51.7 &65.8 &62.1  &80.0 &38.2 &52.0 &66.1     \\
    RAU\cite{Noh2015}      &63.3   &81.9 &39.0 &53.0  &67.7  &63.2   &81.7  &38.2  &52.8  &67.3    \\
    DAN\cite{nam2016dan}              &64.3   &83.0 &39.1 &53.9  &{\textbf{69.1}}  &64.2   &82.8  &38.1  &54.0  &69.0    \\
    MCB+Att\cite{fukui2016multimodal}    &64.2   &82.2 &37.7 &54.8 &-  &- &- &- &- &-  \\
    MCB+Att+GloVe\cite{fukui2016multimodal}    &64.7   &82.5 &37.6 &56.6 &-  &- &- &- &- &-  \\
    MLB \cite{kim2016hadamard}          &65.1   &84.1 &38.2 &54.9 &-    &-  &- &- &- &-  \\
    MFB+CoAtt+GloVe \cite{Yu_2017_ICCV}  &65.9   &84.0 &{\textbf{39.8}} &56.2 &-    &65.8  &83.8 &{\textbf{38.9}} &56.3 &-  \\
    SVA \cite{Zhu_2017_ICCV}          &{\textbf{66.0}}   &{\textbf{84.3}} &39.3 &56.4 &-    &65.9  &{\textbf{84.4}} &{\textbf{38.9}} &55.9 &-  \\
    \midrule
    VKMN(ours)             &{\textbf{66.0}}  &83.7 &37.9 &{\textbf{57.0}} &{\textbf{69.1}}  &{\textbf{66.1}} &84.1 &38.1 &{\textbf{56.9}}   &{\textbf{69.1}}  \\
    \bottomrule
  \end{tabular}
  }
  \vspace{-2ex}
  \caption{Results comparison with state-of-the-art methods on the VQA v1.0 dataset. ``-" indicates the result is not available.
    ``Att" indicates some attention mechanism is used. ``CoAtt" means co-attention.
    ``GloVec" indicates that the word embedding method \cite{Pennington} is adopted.
    Here all the reported results are obtained with a \textbf{single model} without model ensemble.
    For each answer-type, the best result is \textbf{bolded}.
}
  \label{Results-table}
  \vspace{-3ex}
\end{table*}

\subsection{Implementation Details}
We implement our model using the Torch framework with RNN packages from \cite{Leonard2015}.
In the input module, we followed MLB \cite{kim2016hadamard} and used ResNet-152\cite{he2016deep} as the backbone network for visual feature extraction.
The MLB attention feature vector $\mathbf{u}$ is of 2,400 dimension, and the question embedding vector $\mathbf{t}$ is also projected to the same dimension.
The joint visual and knowledge embedding module output 300-dimensional features by $\Phi(e)$.
We tried different embedding methods, such as BOW (GloVe) \cite{Pennington}, and TransE \cite{bordes2013TransE}, and finally pick TransE \cite{bordes2013TransE}.
In the memory module, we set the memory-slot number to 8. If the extracted subgraph size is less than 8, we fill empty slots with zeros.

\subsection{Ablation Studies}
There are several configurations of VKMN which may impact the final accuracy, including knowledge encoding method $\Phi(\cdot)$,
blind model with text feature only, and the triple replicated memory blocks, etc.
We examine the impact of these configurations with evaluation on the Real Image Open-Ended task using the \emph{test-dev} split of the VQA v1.0 dataset.
To avoid interaction between these factors, we adopt the variable-controlling approach to ease our study.
We take our designed model as baseline, which uses TransE as knowledge encoding,
visual attentive feature $\mathbf{u}$ plus knowledge embedding $\mathbf{e}$ to build the keys and values in VKMN, and the triple replicated memory-blocks as discussed in Section~\ref{sec:vkmn} to avoid the ambiguity from knowledge extraction from question.

We evaluate the contribution of each component by either removing it or replacing it with another commonly used component, with all the other components fixed.
\begin{itemize}
\setlength\topsep{1pt}
\setlength\parskip{1pt}
\setlength\itemsep{1pt}
\item[(1)] We replace TransE based knowledge encoding in our design with BOW encoding \cite{Pennington}.
    Results show that TransE performs better than BOW, especially on the ``other" answer-type (57.0 vs 56.1).
\item[(2)] We remove the visual input and design a \textit{blind model} without any visual cues, so that key/value are only text features, i.e., replacing Eq.\ref{eq:embed} with
    \[
    \setlength\abovedisplayskip{1ex}
        \mathbf{x} = \Psi(\mathbf{e}, \mathbf{t}) = \sigma(W_e \Phi(\mathbf{e}) )\odot\sigma(W_t \mathbf{t}),
    \setlength\belowdisplayskip{1ex}
    \]
    and directly using $\mathbf{t}$ as the query. This comparison verifies the importance of visual attentions in VKMN.
\item[(3)] We directly use the joint embedding $\mathbf{q} = \mathbf{t} \odot \mathbf{u}$ for answer prediction without using memory blocks.
    This is actually the result by MLB.
    It is obvious that VKMN outperforms MLB, especially on the ``other" answer-type (57.0 vs 54.9).
    The difference is \textit{significant} according to the paired $t$-test.
    This study verifies the effectiveness of the proposed memory network module.
\item[(4)] We disable the triple replication mechanism in memory networks, while only use $s$ and $r$ to build the key and $t$ as the value.
This study shows that the triple replication mechanism is important to avoid
the ambiguity in visual-question pair, especially on the ``other" answer-type (57.0 vs 53.9).
\end{itemize}
\autoref{tab:ablation} lists the detailed results of the four cases in comparison to our designed model (baseline) on different question categories.
This ablation study verifies the effectiveness of our design choice.

\begin{figure*}[]
\centering
\includegraphics[width=0.9\linewidth]{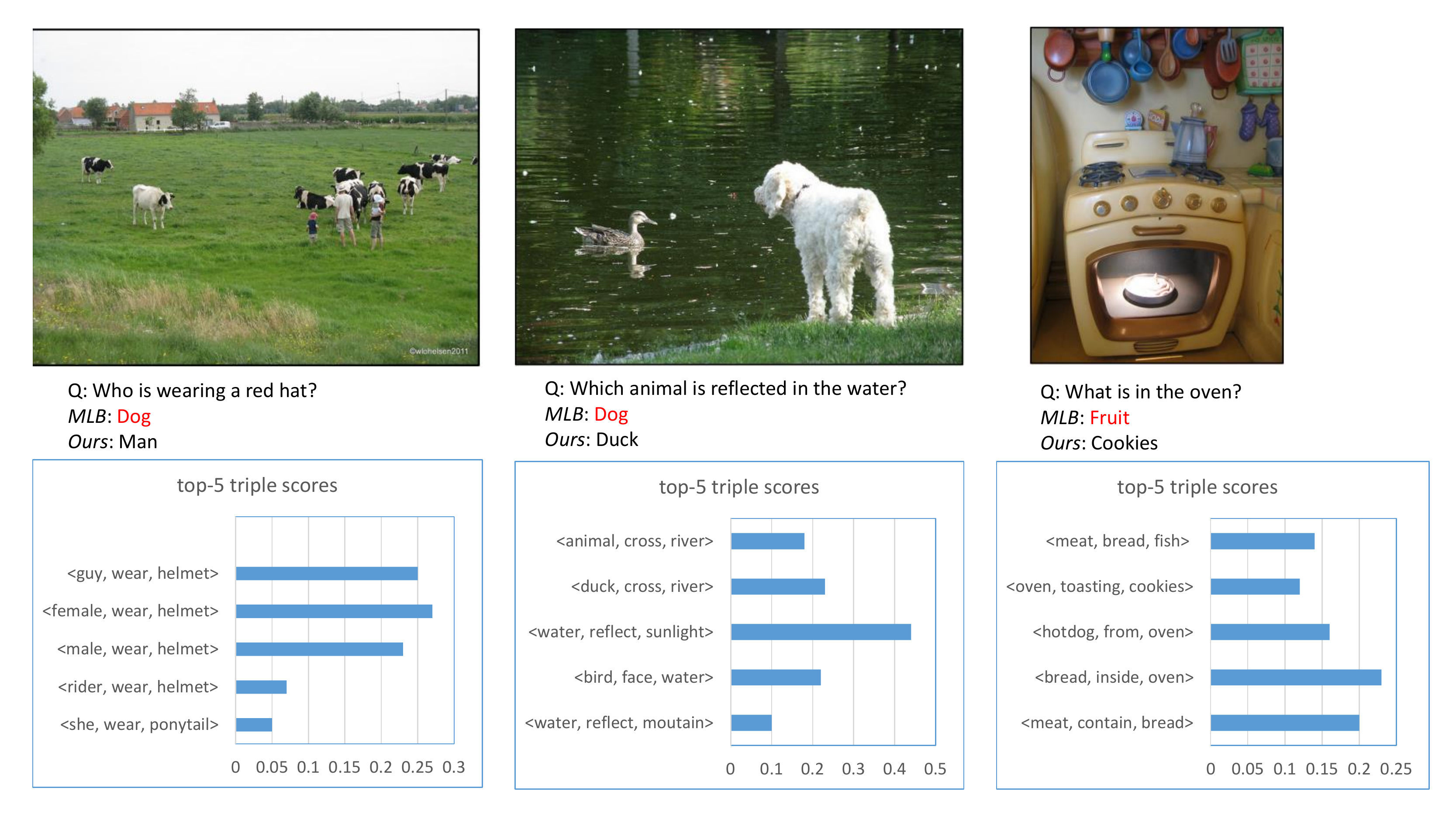}
 \vspace{-1ex}
\caption{Examples of the predicted answers and top-5 knowledge triple scores for given query images and questions. The predicted answers by MLB (used as our attention module) is also given as comparison.}
\label{fig:examples}
  \vspace{-3ex}
\end{figure*}

\subsection{Benchmarking Results}
\begin{figure}[]
\centering
\includegraphics[width=0.95\linewidth]{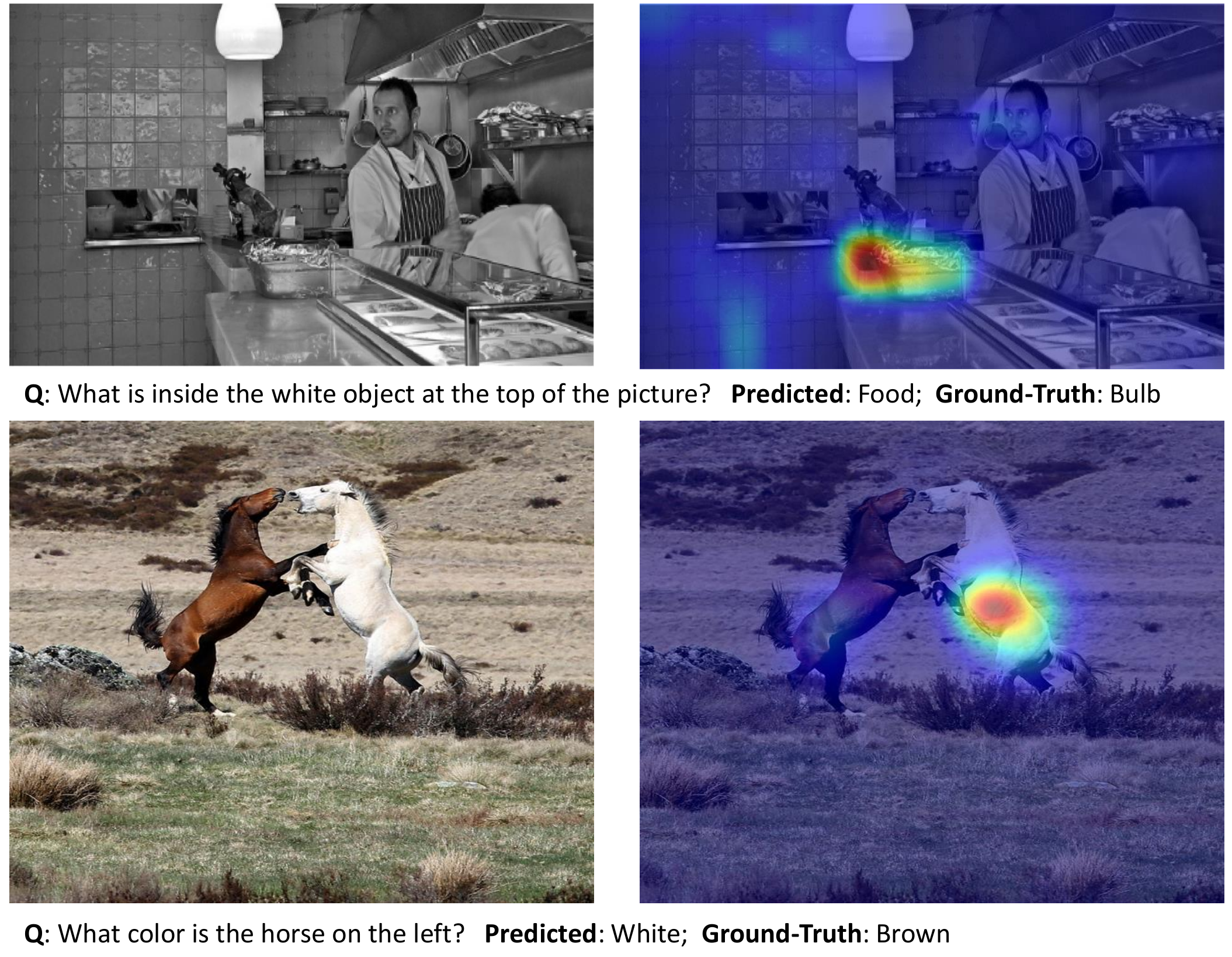}
\vspace{-1ex}
\caption{Some failure cases with attention maps. Both cases are due to wrong attention regions.}
\label{fig:fail}
\vspace{-2ex}
\end{figure}

We further list the full benchmark results on both test-dev and test-standard dataset of the VQA v1.0 dataset in \autoref{Results-table}.
For easy and fair comparison, we also list the results by state-of-the-art methods with single model.
It shows that VKMN outperforms state-of-the-art results on both \emph{Open-Ended} and \emph{Multiple-choice} tasks, and especially for the ``other" answer-type,
which proves that VKMN is effective on incorporating external knowledge for answering 6W questions \textit{(what, where, when, who, why, and how)}.
\autoref{fig:examples} further illustrates some quantitative examples, in comparison to the state-of-the-art method MLB.
Below each example, we also show the confidence score of top-5 knowledge triples according to Eq.\ref{eq:score}.
It is obvious that the VKMN model could attend to highly relevant knowledge triples.
Note that although some top-1 triple is relevant but not quite accurate (due to appearance frequency in training-set),
the final decision is based on softmax classifier (Eq.\ref{eq:smax}) with weight averaging knowledge representation (Eq.\ref{eq:pav}),
which tends to produce right answers.
Besides, we show some failure cases in \autoref{fig:fail} along with the MLB attention maps.
These cases are related to spatial relationship reasoning, in which MLB does not get correct attention regions. Problem may be alleviated when
resorting to some advanced attention mechanism such structure attention \cite{Zhu_2017_ICCV}.

We further evaluated our method on VQA v2.0 dataset, which is about twice larger than VQA v1.0 on question number, and much more difficult than VQA v1.0.
\autoref{tab:vqa2} listed the comparison results on test-standard set.
Single VKMN model achieves 64.36 overall accuracy, which is much better than MCB and MLB (ours retrained model and result submitted by the DCD-ZJU team), especially on the ``other" answer-type.
Our ensemble result is based on the simplest snapshot ensemble \cite{huang2017snapshot} of our VKMN model and the SVA model \cite{Zhu_2017_ICCV}
as we observed that VKMN and SVA are complementary on different answer-types.
Although the top-3 solutions in the VQA v2.0 challenge show better accuracy than our VKMN, they heavily rely on model ensemble,
while their best results are obtained by  ensembling dozens or even hundreds of models.
Even for the first place solution \cite{teney2017tips},
their best single model result is still much worse than that of our single VKMN model, especially on the ``other" answer-type,
when leaving out the exhaustive bottom-up attention from the object detection results.

\section{Conclusion}
In this paper, we present the Visual Knowledge Memory Network (VKMN) method as an efficient way to leverage pre-built visual knowledge base for accurate visual question answering.
Experiments show that VKMN achieves promising results on VQA v1.0 and v2.0 benchmarks,
and outperforms state-of-the-art methods on the knowledge-reasoning related questions (i.e., the ``other" answer-type in both benchmarks).

\begin{table}[]
  \centering
  \small
  \resizebox{\linewidth}{!}{
  \begin{tabular}{cccccc}
    \hline
    \multicolumn{1}{c}{Model} &\multicolumn{1}{c}{All} &\multicolumn{1}{c}{Y/N} &\multicolumn{1}{c}{Num} & \multicolumn{1}{c}{Other}\\
    \hline
    Adelaide-Teney-MSR (30 ensemble)\cite{teney2017tips}   &69.13   &85.54 &47.45 &59.82 \\
    Adelaide-Teney-MSR (best single + bottom-up)\cite{teney2017tips}   &65.67   &82.20 &43.90 &56.26 \\
    Adelaide-Teney-MSR (best single)\cite{teney2017tips}   &62.27   &79.32 &39.77 &52.59 \\
    DLAIT \cite{vqa2leadboard}                  &68.22 &83.17 &46.66   &60.15  \\
    HDU-USYD-UNCC \cite{vqa2leadboard}          &68.09  &84.5   &45.39  &59.01 \\
    \hline
    \hline
     MCB                    &62.27  &78.82  &38.28  &53.36\\
     MLB (by DCD-ZJU)\cite{vqa2leadboard}          &62.54  &79.85  &38.64  &52.95 \\
     MLB (our retrained single model, +VG)             &63.50  &77.73   &39.12  &56.98 \\
     SVA (our retrained single model) \cite{Zhu_2017_ICCV} &64.55 &80.76 & 42.59 & 55.73 \\
    \hline
    \hline
    VKMN (Single)             &64.36  &83.70 &37.90 &57.79\\
    VKMN (Ensemble)           &66.67  &82.88  &43.17  &57.95\\
    \hline
  \end{tabular}
  }
  \vspace{-2ex}
  \caption{Results on VQA v2.0 test-standard set. The first group of results are from top-3 solutions in the VQA v2.0 challenge.
  The second group of results are from some attention based methods related to us.
  For our retrained MLB model on VQA v2.0, we include the external visual genome (VG) data. The final group of results are from our VKMN. }
  \label{tab:vqa2}
 \vspace{-3ex}
\end{table}

\vspace{1pt}
\noindent\textbf{Acknowledgements} Thanks Zhiqiang Shen for the help of preparing some illustrations for our early submissions.

\begin{table*}[]
  \centering
  \small
  \begin{tabular}{cccc|cccc|cccc}
    \hline
    \multirow{2}*{Group} & \multirow{2}*{Min area} & \multirow{2}*{Max area} & \multirow{2}*{\#Questions} & \multicolumn{4}{c}{MLB} & \multicolumn{4}{c}{VKMN}  \\
                            &                         &                         &                           & Other & Number & Y./N. & All & Other & Number & Y./N. & All \\
    \hline
    1 & 44.320 & 4955.855 & 7829 & 58.436 & 43.210 & 59.122 & 56.927 & 69.685 & 48.148 & 60.494 & 63.512 \\
    2 & 4955.855 & 16753.460 & 7829 & 56.653 & 51.303 & 59.122 & 57.065 & 69.136 & 55.556 & 59.808 & 63.786 \\
    3 & 16753.460 &  36452.097  & 7829  &  56.927 & 45.679 & 60.357 & 57.065 & 68.176 & 50.206 & 61.180 & 63.237 \\
    4 & 36452.097 &  65408.207 &  7829  & 53.772 & 47.188 & 66.118 & 58.162 & 66.804 & 51.440 & 67.078 & 65.158 \\
    5 & 65408.207 &  370630.380 & 7830  &  55.823 & 36.895 & 67.618 & 58.566 & 66.521 & 64.327 & 68.030 & 64.327 \\
    \hline
  \end{tabular}
  \caption{Results on different object-size groups.}\label{tab:size}
\end{table*}
\subsection*{Appendix: Influence of Target Object Size}
We study the influence of target object size by comparing our VKMN and MLB performance on VQA 2017 test-dev set at different object-size groups.
As the dataset does not have the target object size information available to public, we ran state-of-the-art object detector, Cascade R-CNN~\cite{cai18cascadercnn}, on the test-dev set, where the detector has 0.427 AP on COCO 2017 challenge.
This study is thus not very precise, but is still able to reveal some trends.

We preserve the question where there is at least one detected object's name is contained in the question or VKMN's answer, which leaves 39146 and  out of 107394 questions.
We take the matched object name as the target, and further divide questions into 5 equal-size groups according to target object size.
For one target with multiple instances in one image, we take the average size as the target size.
Finally, we upload the results to the test server to get the accuracies of VKMN and MLB at different object-size groups, and show the results in \autoref{tab:size}.
From the table, we could draw the following conclusions:
(1) For both MLB and VKMN, larger targets show higher overall accuracy than smaller targets, especially on the ``Yes/No" answer-type;
(2) On the ``other" answer-type, different object-size groups have almost consistent performance for both MLB and VKMN;
(3) VKMN outperforms MLB on the  ``other" answer-type significantly on all the object-size groups.

{\small
\bibliographystyle{ieee}
\bibliography{ref_vkmn}
}

\end{document}